\documentclass[10pt,twocolumn,letterpaper]{article}

\usepackage{cvpr}
\usepackage{times}
\usepackage{epsfig}
\usepackage{graphicx}
\usepackage{amsmath}
\usepackage{amssymb}
\usepackage[caption=false, font=footnotesize]{subfig}
\usepackage{caption}

\usepackage[breaklinks=true,bookmarks=false]{hyperref}

\cvprfinalcopy 

\def\cvprPaperID{****} 

\begin{document}

\title{AdapterNet - learning input transformation for domain adaptation}

\author{ Alon Hazan, Yoel Shoshan, Vadim Ratner\\
IBM Research - Haifa\\
{\tt\small }
\and
Daniel Khapun, Roy Aladjem\\
Technion - Haifa\\
{\tt\small }
}

\maketitle

\begin{abstract}
   Deep neural networks have demonstrated impressive performance in various machine learning
   tasks. However, they are notoriously sensitive to changes in data distribution. Often, even a slight change in the distribution can lead to drastic performance reduction. Artificially augmenting the data may help to some extent, but in most cases, fails to achieve model invariance to the data distribution. 
   Some examples where this sub-class of domain adaptation can be valuable are various imaging modalities such as thermal imaging, X-ray, ultrasound, and MRI, where changes in acquisition parameters or acquisition device manufacturer will result in different representation of the same input. Our work shows that standard fine-tuning fails to adapt the model in certain important cases. We propose a novel method of adapting to a new data source, and demonstrate near perfect adaptation on a customized ImageNet benchmark. Moreover, our method does not require any samples from the original data set, it is completely explainable and can be tailored to the task. 
\end{abstract}

\section{Introduction}

When training neural networks for a certain task on a specific dataset, it is impossible to guarantee adequate performance on data that has even a slightly different distribution. This challenge gives rise to the field of domain adaptation, which attempts to alter a source domain to bring its distribution closer to that of the target.

A common practice in such cases is to use fine-tuning. Fine-tuning initializes the network to previously learned weights and then retrains it on the new data set, usually with a lower learning rate. The training is almost always limited to the last few layers of the network, while keeping the rest of the network fixed.
The fine-tuning technique is widely adopted due to the simplicity of the process, and the performance achieved. In practice, sometimes the dataset used to train the original model is not available due to legal issues, privacy concerns, or logistics; fine-tuning does not require the original data. 
In the related work section, we discuss additional domain adaptation techniques, for example, techniques to achieve domain representation invariance.

In this paper we propose a method that is just as simple to use and, in some cases, significantly outperforms the fine-tuning approach. Our novel method deviates from the current trend of retraining a subset of the pre-trained model weights. Instead, we introduce an additional network that transforms the input before it is passed to the original pre-trained model, while keeping the original model intact. We demonstrate our method on a modified version of ImageNet, in which images are transformed to simulate different imaging sources. We show that our method manages to overcome the data distribution gap, where both naive inference and fine-tuning approaches fail.

One might think that adding a neural network before the original model is equivalent to training the bottom layers of the DNN, but there are two major differences. First, the added network can be detached from the original model after training, allowing us to observe the output image after adaptation to make sure it is not destructive and provide expandability. Second, unlike the bottom layers, which are pre-determined, by adding a network we have full control of its architecture. This allows us to control its capacity and tailor the network to the expected task e.g., allow only color change.

Our approach is not limited to classification tasks or to image related targets.

\section{Method}
\begin{figure}
	\begin{center}
		\includegraphics[width=1.0\linewidth]{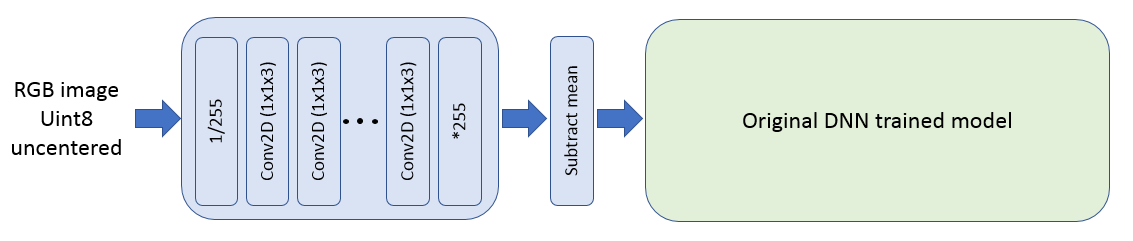}
	\end{center}
	\caption{AdapterNet architecture, initialized to be an identity function. We subtract the mean in order to include the expected preprocessing of the pre-trained model. The original DNN acts as a big loss function in training the small AdapterNet}
	\label{fig:adapter_net}
\end{figure}

We propose a method to create and train an adapter network to transform data from the new data source to the original data distribution. For the adapter, which we call AdapterNet, we use a shallow convolutional neural network to enforce a low capacity. However, any model architecture can be used, depending on the expected complexity of the required transformation. Figure \ref{fig:adapter_net} illustrates the AdapterNet architecture. 

To train the adapter, we treat the original pre-trained network together with the original loss function as a single large loss function. In other words, we freeze all the weights of the original network and add the adapter network before the input. 
We then train the adapter with supervision samples from the new data set. Since the adapter is a shallow network, training it requires fewer samples than is required to train a full new network and as such, it trains quickly. The adapter learns by doing the exact same task that was used to train the original network (e.g., classification). However, since it is connected to a pre-trained network and its capacity is limited due to the small number of parameters, we expected it to learn the representation transformation. 

By experimenting with different initialization schemes, we discovered that initialization of the adapter weights is crucial to the success of the training. When we used standard random initialization schemes such as Glorot \cite{Glorot2010} or He \cite{He2015}, the model completely failed to learn. Instead, we initialized the adapter to begin as an identity function, which means that before training starts the adapter function is effectively F(x)=x. We achieved this by using Relu \cite{Nair2010} activation and providing inputs that are greater or equal to zero, to guarantee the identity function starting point. We transformed the input data into the range [0,1]. Our suggested convolutional layers are spatially 1x1 with 3 channels; therefore, they allow information mixture between channels. However, in the identity initialization, we enforced zero weights for inter-channel connections and unit weights for intra-channel connections; this means that initially the adapter was an identity function, regardless of the number of layers we chose to use.

\section{Related Work}\label{related_work}

A wide range of domain adaptation methods attempt to overcome the gap between source and target data distribution, which often causes a significant performance degradation. Most of these methods focus on either unsupervised domain adaptation or semi-supervised domain adaptation. We are not aware of any method that simultaneously combines the following four requirements: 1) doesn't require any samples from the original data set. 2) Can be detached and used as standalone image adapter. 3) Can be tailored architecturally to a specific task. 4) Requires a small amount of samples from the new data set.

In AdaBN \cite{Li2018}, batch normalization statistics parameters are updated to better match the target domain. This differs greatly from our method, since our method does not assume any normalization method and is orthogonal to it. In addition, since our method is not intrusive to the original network, we can separate our learned transformation function from the full network, which helps us inspect the learned transformation effect.
Work by \cite{Yao2015} simultaneously minimizes the classification error while also restricting a similarity term on unlabeled target examples. It also involves jointly training on source and target domains. Our approach does not require any modification of the originally desired loss function. Moreover, it does not require joint training, as we train only the adapter module while keeping the pre-trained source domain classifier fixed.
Work by \cite{Zhou2017} attempts to project samples from both source and target domains into a common latent space by minimizing the ratio of within-class distance to between-class distance. Our approach does not require jointly learning a feature extractor and/or classifier with samples from both source and destination domains available at training time. In addition, our method does not require training the entire feature extractor and/or classifier. Instead, we learn a relatively modest initial adapter block, which is significantly faster to train due to the small number of parameters.

Recently, domain adaptation methods inspired by generative adversarial networks (GAN) \cite{Goodfellow2014} have gained popularity. For example, \cite{Ganin2015} added a domain discriminator to enforce domain invariance of the extracted features. It can be seen as two agents competing in a minmax game. One agent is trying to minimize classification error and maximize discriminator domain confusion. Another agent is trying to minimize discriminator domain confusion. Work by \cite{Wulfmeier2017} use this technique to overcome different outdoor lighting conditions between source and destination domains. \cite{Volpi2018} add feature augmentation on top of the domain discriminator. GAN inspired methods differ from our approach, as our approach which does not involve a discriminator and/or generator. GANs are also notoriously hard to train. Despite the development of techniques to improve the training process, gradient explosion and mode collapse are just a few of the many common practical issues faced when trying to train a GAN-based model.

Recent surveys \cite{Mei2018}, \cite{Cheplygina2018}, \cite{Csurka2017a} may provide additional background on domain adaptation and transfer learning.

\section{Creating a benchmark}
To test the algorithm, we customized the ImageNet \cite{Russakovsky2015} dataset in version ILSVRC2012. We considered the original train set of ImageNet (1,281,167 images) as inaccessible; we could only use the model that was already trained with it as a given. Instead, we used the original ImageNet validation set of 50K images in the following way:
We took IDs 1 - 40,000 as the new "train set", IDs 40,001 - 45,000 as the new "validation set" and
IDs 45,001 - 50,000 as the new "test set".  

We applied transformations on all the 50K images to simulate different representations of the same domain. Our goal was to show that our algorithm is capable of transforming the input in a way that together with the original model, performs almost as well as the performance on the original data set. 

We applied two different transformations on the data; each transformation creates a simulated new camera. We use it to train and evaluate our adaptation method and compare it to traditional fine-tuning which failed in this task. We detail the transformations in Section \ref{simutalted_cameras}

\section{Simulated cameras}\label{simutalted_cameras}

To simulate different image sources, we created transformation functions and applied them on ImageNet images. Ideally, the new data source will contain the same amount of information as the original, but will be represented differently. Thus, we used transformations that are lossless. In practice, due to numerical quantization, the transformations we chose do lose some information. Nonetheless, our algorithm was able to overcome this obstacle.
Our method is not expected to perform well on all domain adaptation tasks, it is specifically designed to address cases were the source and target differ in parameters of acquisition and not in essence. When selecting transformations, we made sure to include non-linear transformations, because such data distribution changes are common in several practical scenarios such as medical imaging, thermal devices, and more. It is worth noting that image augmentation during training is considered a standard, however we know that it still can't cover all variations between two data sets, especially if the during training of the original model, we don't have information about the future data set, or can't describe the differences between them in the form of augmentation.

\subsubsection{Color rotation}

We performed color space rotation to simulate a camera that captures colors substantially differently from the original camera. To rotate the color space without affecting the luminosity, we performed the rotation in the CIELAB color space. We looked at each pixel’s a,b channels as a 2D vector that we rotate at an angle of 150 degrees. Doing so reduces the classification performance of the original model by 34.6\% (from top-1 score of 0.6546 to 0.4282 on the color-rotated images). To choose the rotation angle for our simulation, we evaluated the original classifier on various angles and selected the worst performing value. Figure \ref{fig:lab_graph} shows the effect of color rotation on the top-1 score. Figure \ref{fig:edge} shows visual examples. Note that rotation of the vector can result in invalid values (e.g., b>128). In this case, they are trimmed so there is some loss of information in this transformation.

\begin{center}
		\includegraphics[width=1.0\linewidth]{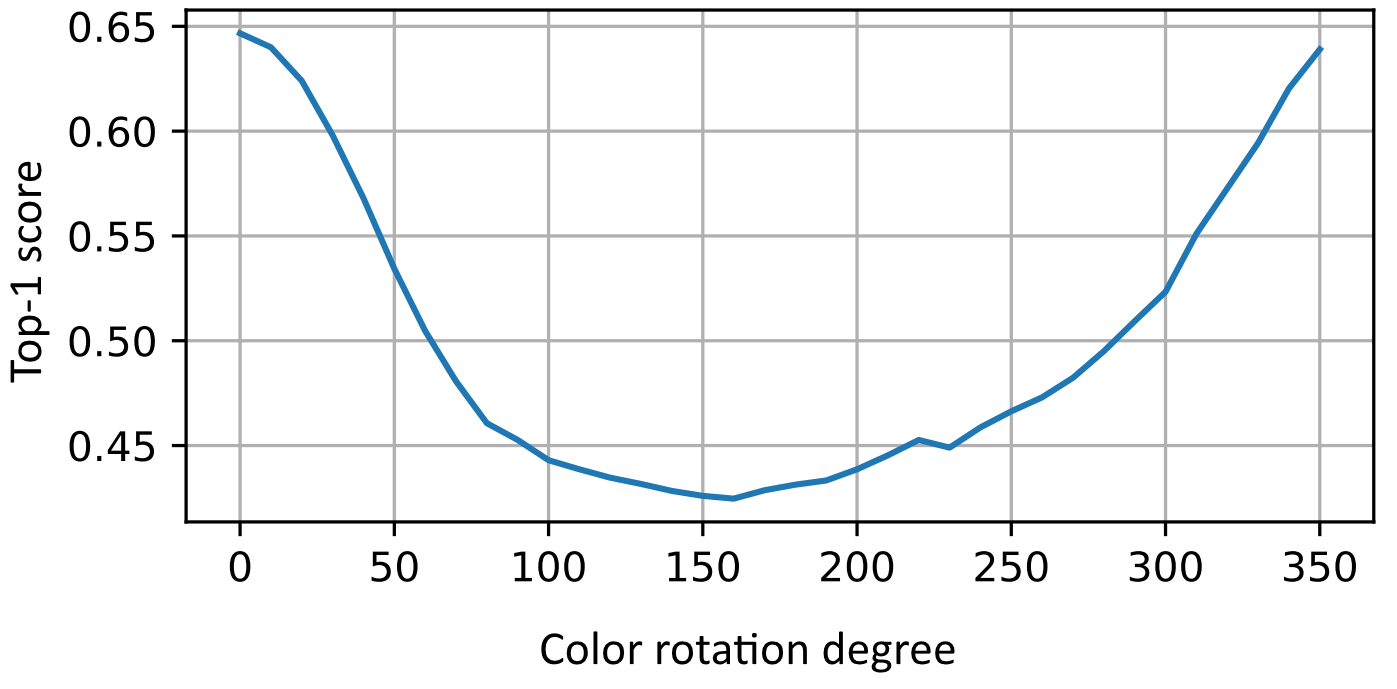}
		\captionof{figure}{VGG16 performance after various color rotation degrees. This plot was performed on a small subset of 2000 images to get qualitative results for the effect of color rotation on classifier performance}\label{fig:lab_graph}
\end{center}
\begin{center}
	\includegraphics[width=.8\columnwidth]{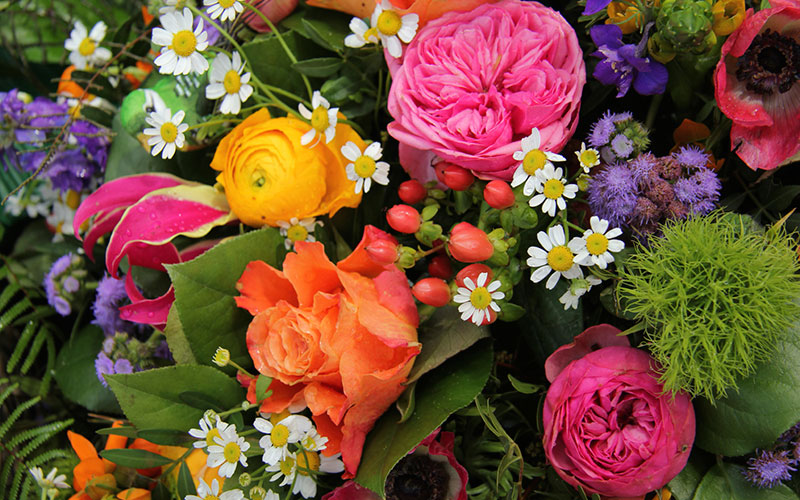}
	\includegraphics[width=.49\columnwidth]{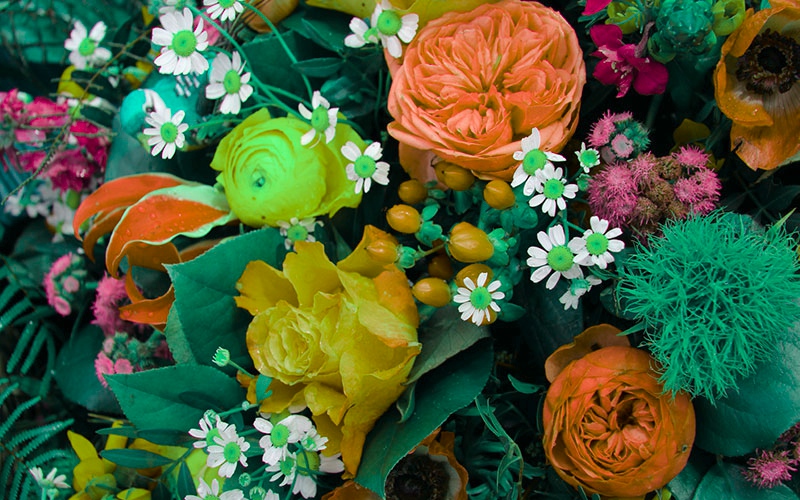}
	\includegraphics[width=.49\columnwidth]{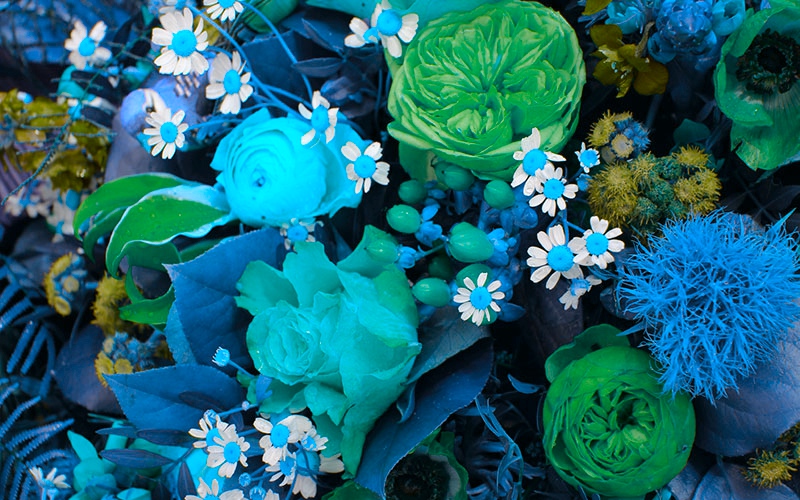}	
	\captionof{figure}{Effect of color rotation on a sample image. Top: Original, Bottom-left: $50^{o}$ rotation, Bottom-right: $150^{o}$ rotation}\label{fig:edge}
\end{center}

\subsubsection{Power function}
We tested a non-linear transformation on the luminosity as well as some color manipulation. We raised the RGB values to the powers  $\widetilde{R} = R^{0.2}, \widetilde{G} = G^{0.3}, \widetilde{B} = B^{0.4}$. Doing so lowers the top-1 score by 33.4\% (from 0.6546 to 0.4356 on the transformed images). This transformation is similar to the gamma factor, which tends to be different across imaging devices. We applied it on the color channels in the range [0,1], thus the luminosity remains in the same valid range. We chose different power coefficients for each channel to introduce some color change as well as gamma. Due to uint8 discretization, the amount of pixel values after the transformation of (for example) power of 0.2 is 120 out of 256; so there is a loss of information in practice. Nonetheless, the AdapterNet was able to learn the adaptation. Figure \ref{fig:power} shows a visual example of the effect of the power function.

\begin{center}
	\includegraphics[width=0.495 \columnwidth]{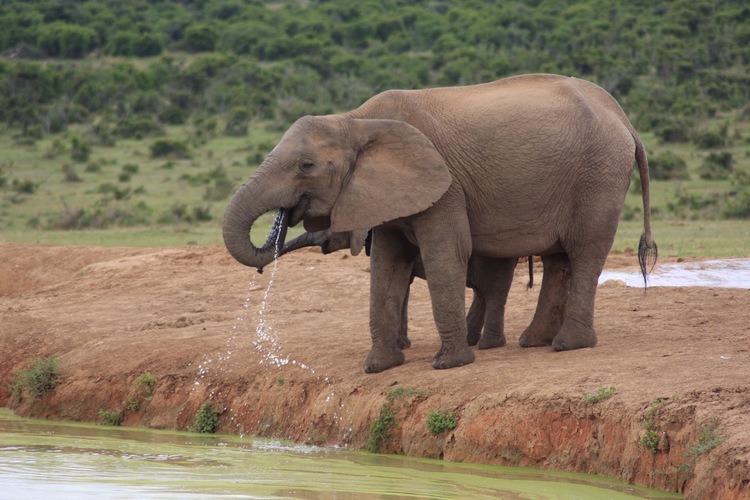}
	\includegraphics[width=0.495 \columnwidth]{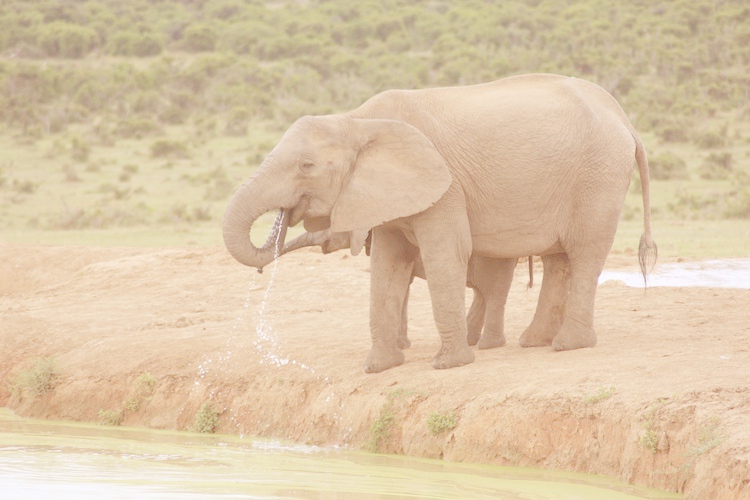} 
	\captionof{figure}{Effect of the power function on a sample image. Top image is the original and bottom image}\label{fig:power}
\end{center}

\subsection{Results}

We tested our method against the traditional fine-tuning approach. The top-1 score of the pre-trained model (VGG16 \cite{Simonyan2015}) on our newly defined test set (original, untransformed) is 0.6546, which serves as the desired goal for performance when inferring on transformed images. 

After applying color rotation, the top-1 score of the pre-trained model dropped by 34.6\% to 0.4282. On a second scenario, applying the power function, it dropped by 33.4\% to 0.4356. We measured how well each method can adapt to the new inputs and repair the performance drop.

First, we experimented with fine-tuning by training only the last trainable layer. We noticed some improvement in the performance for the color rotation, and no improvement in the power function transformation. We then tested fine-tuning on the last two trainable layers,which showed significant improvement for both filters. However, there was clear over-fitting after a few epochs, no matter what optimizer or learning rate we tried. At this point it was a likely speculation that any more trainable layers would just cause more over-fitting. To strengthen that notion, we did a third fine-tuning experiment in which the last three trainable layers are unfrozen. We observed that the over-fitting occurred after even fewer epochs, and the model failed to adapt to the new source.
With our AdapterNet with 5 layers, the results show near perfect adaptation performance for both tested transformations. 

\begin{table*}
	\label{finetune_result}
	\begin{center}
		\begin{tabular}{|l|c|c|}
			\hline
			Method & Color rotation score ([\%] Drop) & Power function score ([\%] Drop) \\
			\hline\hline
			Pure inference                & 0.4282 (34.6\%) & 0.4356 (33.4\%) \\
			Fine-tuning last layer        & 0.4821 (26.3\%) & 0.4362 (33.4\%) \\
			Fine-tuning last two layers   & 0.5215 (20.3\%) & 0.5134 (21.6\%) \\
			Fine-tuning last three layers & 0.4343 (33.6\%) & 0.4322 (34\%) \\
			AdapterNet                    & 0.631 (3.6\%)  & 0.6188 (5.5\%) \\
			\hline
		\end{tabular}
	\end{center}
	\caption{The table shows the performance after each of the image transformations. 
		The performance metric is the top-1 score for classification of the ImageNet data set. 
		We also note in parentheses the percentage of performance drop, relative to the original top-1 score of 0.6546 achieved by the pre-trained model on our clean test set.}
\end{table*}

The AdapterNet has the advantage of being a separate element from the pre-trained network, so we can examine it independently. After training the AdapterNet to adapt to the image transformations, we looked at the images at the output of the adapter before they are input to the pre-trained network. We expected to see the images restored to the original appearance. The images in Figures \ref{fig:final_lab} and \ref{fig:final_power} resemble the original; although the colors don't always exactly match the original, they still look realistic, and perhaps they did not contribute to the classification task. We noticed that the images had more contrast after the adapter. Perhaps in some ways, they were better than the original; this suggests that the AdapterNet may also be used to enhance images prior to inference even from the same domain. However, testing this is beyond the scope of this work and may be addressed in future work.

	\begin{center}
		\includegraphics[width=.49\columnwidth]{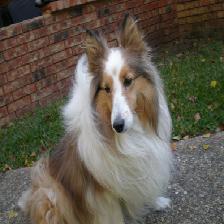}
		\includegraphics[width=.49\columnwidth]{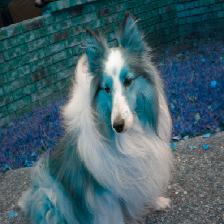}
		\includegraphics[width=0.8\columnwidth]{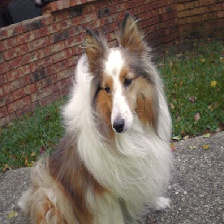}
		\captionof{figure}{AdapterNet's effect on color rotation. Top-left: Original, Top-right: Transformed, Bottom: After Adapter}\label{fig:final_lab}
	\end{center}
	
\newpage

	\begin{center}
		\includegraphics[width=.49\columnwidth]{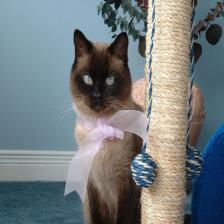}
		\includegraphics[width=.49\columnwidth]{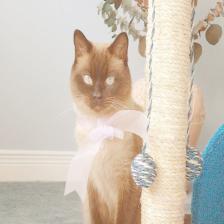}
		\includegraphics[width=.8\columnwidth]{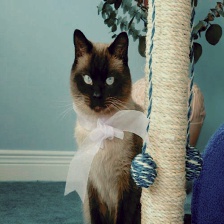}
		\captionof{figure}{AdapterNet's effect on power function. Top-left: Original, Top-right: Transformed, Bottom: After Adapter}\label{fig:final_power}
	\end{center}

\subsubsection*{Discussion}

We demonstrated a simple way to implement a method for domain adaptation in cases where there is data from a new distribution, but with the same task. We also showed that it this method performs significantly better than fine-tuning, without adding much complexity or many new hyper-parameters. In addition, it does not require any of the data that was used to train the original model.
This method, as we refer to as AdpaterNet, can also work orthogonally to fine-tuning and batch renormalization.

The AdapterNet can be taken as a standalone network to convert samples from one distribution to another. We showed image samples and observed that in some ways, the transformed images appear visually better than the original. We did not formally evaluate this effect, since it is beyond the scope of this work. We note it as possible future work to see if it can serve as an image enhancement mechanism. 

We also noted that the transformations we used to simulate different imaging sources were lossy in practice, in some case reducing the range of pixel values from 256 to 120, for example. The adapter was able to overcome the information loss. This suggests that there may be a way to compress images without affecting performance, if so, it can answer some Big-Data storage cost issues.

For future work, we are considering implementing the same AdapterNet concept in different places inside the pre-trained network, rather than before it. Using the identity initialization, we speculate that we can achieve inner layer adaptation, as well.

{\small
\bibliographystyle{ieee}
\bibliography{egbib}

\begin{thebibliography}{10}\itemsep=-1pt

\bibitem{Cheplygina2018}
V.~Cheplygina, M.~de~Bruijne, and J.~{P. W. Pluim}.
\newblock {Not-so-supervised: a survey of semi-supervised, multi-instance, and
  transfer learning in medical image analysis}, 2018.

\bibitem{Csurka2017a}
G.~Csurka.
\newblock {Domain Adaptation for Visual Applications: A Comprehensive Survey}.
\newblock {\em survey}, pages 1--46, 2017.

\bibitem{Ganin2015}
Y.~Ganin and V.~Lempitsky.
\newblock {Unsupervised Domain Adaptation by Backpropagation}.
\newblock In {\em Proceedings of the 32nd International Conference on Machine
  Learning, PMLR}, pages 37:1180--1189, 2015.

\bibitem{Glorot2010}
X.~Glorot and Y.~Bengio.
\newblock {Xavier}.
\newblock {\em Proceedings of the 13th International Conference on Artificial
  Intelligence and Statistics (AISTATS)}, 9:249--256, 2010.

\bibitem{Goodfellow2014}
I.~Goodfellow, J.~Pouget-Abadie, M.~Mirza, B.~Xu, D.~Warde-Farley, S.~Ozair,
  A.~Courville, and Y.~Bengio.
\newblock {Generative Adversarial Nets}.
\newblock {\em Advances in Neural Information Processing Systems 27}, pages
  2672--2680, 2014.

\bibitem{He2015}
K.~He, X.~Zhang, S.~Ren, and J.~Sun.
\newblock {Delving deep into rectifiers: Surpassing human-level performance on
  imagenet classification}.
\newblock In {\em Proceedings of the IEEE International Conference on Computer
  Vision}, volume 2015 Inter, pages 1026--1034, 2015.

\bibitem{Li2018}
Y.~Li, N.~Wang, J.~Shi, X.~Hou, and J.~Liu.
\newblock {Adaptive Batch Normalization for practical domain adaptation}.
\newblock {\em Pattern Recognition}, 80:109--117, 2018.

\bibitem{Mei2018}
W.~Mei and D.~Weihong.
\newblock {Deep Visual Domain Adaptation: A Survey}, 2018.

\bibitem{Nair2010}
H.~Qian.
\newblock {Cooperativity in cellular biochemical processes: noise-enhanced
  sensitivity, fluctuating enzyme, bistability with nonlinear feedback, and
  other mechanisms for sigmoidal responses.}
\newblock {\em Annual review of biophysics}, 41(3):179--204, 2012.

\bibitem{Russakovsky2015}
O.~Russakovsky, J.~Deng, H.~Su, J.~Krause, S.~Satheesh, S.~Ma, Z.~Huang,
  A.~Karpathy, A.~Khosla, M.~Bernstein, A.~C. Berg, and L.~Fei-Fei.
\newblock {ImageNet Large Scale Visual Recognition Challenge}.
\newblock {\em International Journal of Computer Vision}, 115(3):211--252,
  2015.

\bibitem{Simonyan2015}
K.~Simonyan and A.~Zisserman.
\newblock {Very Deep Convolutional Networks for Large-Scale Image Recognition},
  2015.

\bibitem{Volpi2018}
R.~Volpi, P.~Morerio, S.~Savarese, and V.~Murino.
\newblock {Adversarial Feature Augmentation for Unsupervised Domain
  Adaptation}, 2018.

\bibitem{Wulfmeier2017}
M.~Wulfmeier, A.~Bewley, and I.~Posner.
\newblock {Addressing appearance change in outdoor robotics with adversarial
  domain adaptation}.
\newblock In {\em IEEE International Conference on Intelligent Robots and
  Systems}, volume 2017-Septe, pages 1551--1558, 2017.

\bibitem{Yao2015}
T.~Yao, Y.~Pan, C.-w. Ngo, H.~Li, and T.~Mei.
\newblock {Semi-supervised Domain Adaptation with Subspace Learning for Visual
  Recognition}.
\newblock {\em Cvpr}, 2015.

\bibitem{Zhou2017}
X.~Zhou and S.~Prasad.
\newblock {Transformation learning based domain adaptation for robust
  classification of disparate hyperspectral data}.
\newblock In {\em 2017 IEEE International Geoscience and Remote Sensing
  Symposium (IGARSS)}, pages 3640--3643. IEEE, jul 2017.

\end{thebibliography}
}

\end{document}